\documentclass[11pt,a4paper]{article}
\usepackage[T1]{fontenc}
\usepackage[utf8]{inputenc}
\usepackage{lmodern}
\usepackage[left=25mm,right=25mm,top=25mm,bottom=25mm]{geometry}
\usepackage{microtype}
\usepackage{amsmath,amssymb}
\usepackage{array,booktabs,tabularx}
\usepackage{graphicx,xcolor,tikz}
\usetikzlibrary{arrows.meta,calc,positioning}
\usepackage{placeins}
\usepackage[font=small,labelfont=bf,labelsep=period]{caption}
\usepackage[numbers,sort&compress]{natbib}
\usepackage{fancyhdr}
\usepackage{xurl}
\usepackage{hyperref}
\definecolor{ink}{HTML}{223744}
\definecolor{accent}{HTML}{225A68}
\definecolor{rulegray}{HTML}{9EAEB3}
\hypersetup{colorlinks=true,linkcolor=accent,citecolor=accent,urlcolor=accent,
 pdftitle={A Data-Isolated Agent Harness for Analytical Workflows},
 pdfauthor={Fredrik A. Dahl},pdfsubject={LLM-assisted construction and model-free execution of analytical workflows}}
\newcommand{\doi}[1]{\href{https://doi.org/#1}{doi:\nolinkurl{#1}}}
\newcolumntype{Y}{>{\raggedright\arraybackslash}X}

\newcommand{\playerref}{\texttt{\$\{parameters.player\}}}
\newcommand{\stepref}{\texttt{\$\{steps.step\_1.artifact\_id\}}}
\usepackage{fvextra}
\fvset{breaklines=true,breakanywhere=true,breaksymbolleft={},
        breaksymbolright={},breakautoindent=false,breakindent=0pt}
\DefineVerbatimEnvironment{TaskPrompt}{Verbatim}{fontsize=\small,
        frame=single,framerule=0.3pt,framesep=4pt}

\title{\bfseries A Data-Isolated Agent Harness for Analytical Workflows}
\author{Fredrik A. Dahl\\Norwegian Computing Center, Oslo, Norway}
\date{28 November 2026}
\makeatletter
\renewcommand{\@maketitle}{%
  \begin{center}
  {\LARGE\@title\par}\vspace{9pt}
  {\large\@author\par}\vspace{5pt}
  {\normalsize\@date\par}\vspace{7pt}
  \end{center}}
\makeatother

\begin{document}
\maketitle

\begin{abstract}
Externally hosted language models can help develop analyses even when the data and results must remain local. We present an agent harness that separates the construction of an analytical procedure from its execution on private data. One model role specifies the task in dialogue with the user, while another explores tool calls against an empty database created from an approved schema. The host freezes the operations required to produce the selected outputs into a parameterized pipeline. Execution on private data, including reporting and export, requires no model calls. A third model role, without tool access, supports minor revisions to the pipeline. The fixed sequence of operations can be inspected and reused, while SQL and the tools handle data-dependent computation.

We demonstrate the generic harness in a chess case study by supplying a chess database, domain-specific tools exposed through the Model Context Protocol (MCP) and Skills, without modifying the harness itself. Chess was chosen for its analytical complexity, the availability of large public game collections, and its comparatively low privacy risks. Two applications use the author's own games: an analysis of game outcomes stratified by the evaluation of the first pawn-ending position, and the preparation of a dataset for subsequent analysis. The resulting pipelines illustrate how the same construction and replay mechanism supports both statistical analysis and research data preparation.
\end{abstract}

\section{Introduction}
Sensitive records and analytical results may need to remain within an organization's own computing environment even when large language models (LLMs) help develop the analysis. Sending those records or results to an external model service discloses them to that service. Hosting models locally avoids this transmission, but requires operating the inference infrastructure and limits the choice to models available for local deployment. We investigate how externally hosted models can help construct analyses without receiving the private data.

The agent harness described here separates construction of a procedure from execution on its data. The host creates an empty database from approved schema definitions. Through dialogue with the user, LLM1 specifies a canonical task and candidate inputs. LLM2 then explores operations provided through the Model Context Protocol (MCP), using the empty database and a separate workspace. It can correct unsuccessful calls and try alternative sequences. Of the private database, the models see only the disclosed structure; they also receive the task, its inputs, and the tool descriptions and instructions needed for construction.

When LLM2 selects the outputs to deliver, the host retains their producing operations and dependencies. This freezing procedure also identifies the inputs used by the retained steps. The resulting pipeline contains the canonical task and the executable procedure. It can run on the real database with different parameter values, including all reporting and export, without calling a model. LLM3 supports minor changes to defaults or presentation without repeating discovery.

The contribution is a general harness for constructing and reusing analyses under this data separation. We test the generic harness in the chess domain because it combines analytical complexity and substantial game collections with comparatively low privacy stakes. Public collections such as the Lichess Open Database illustrate the availability of data~\cite{lichess_open_database}; the applications in this article use the author's own games. Position selection, player perspectives, and joins between games and positions provide demanding tasks whose calculations can be inspected in detail.

Generating analyses from API descriptions or metadata while keeping records outside model processing has close precedents~\cite{zeng2023flowmind,pawlak2026metadata,beurerkellner2025patterns}. Our focus is the construction mechanism: tool exploration against an empty database, freezing from selected outputs, and subsequent replay and revision. We first describe the harness, then its chess implementation and two applications, before comparing related work and discussing the evidence and scope of the design.

\section{The agent harness}
\label{sec:construction}
\begin{figure}[!htbp]
\centering
\includegraphics[width=0.88\linewidth]{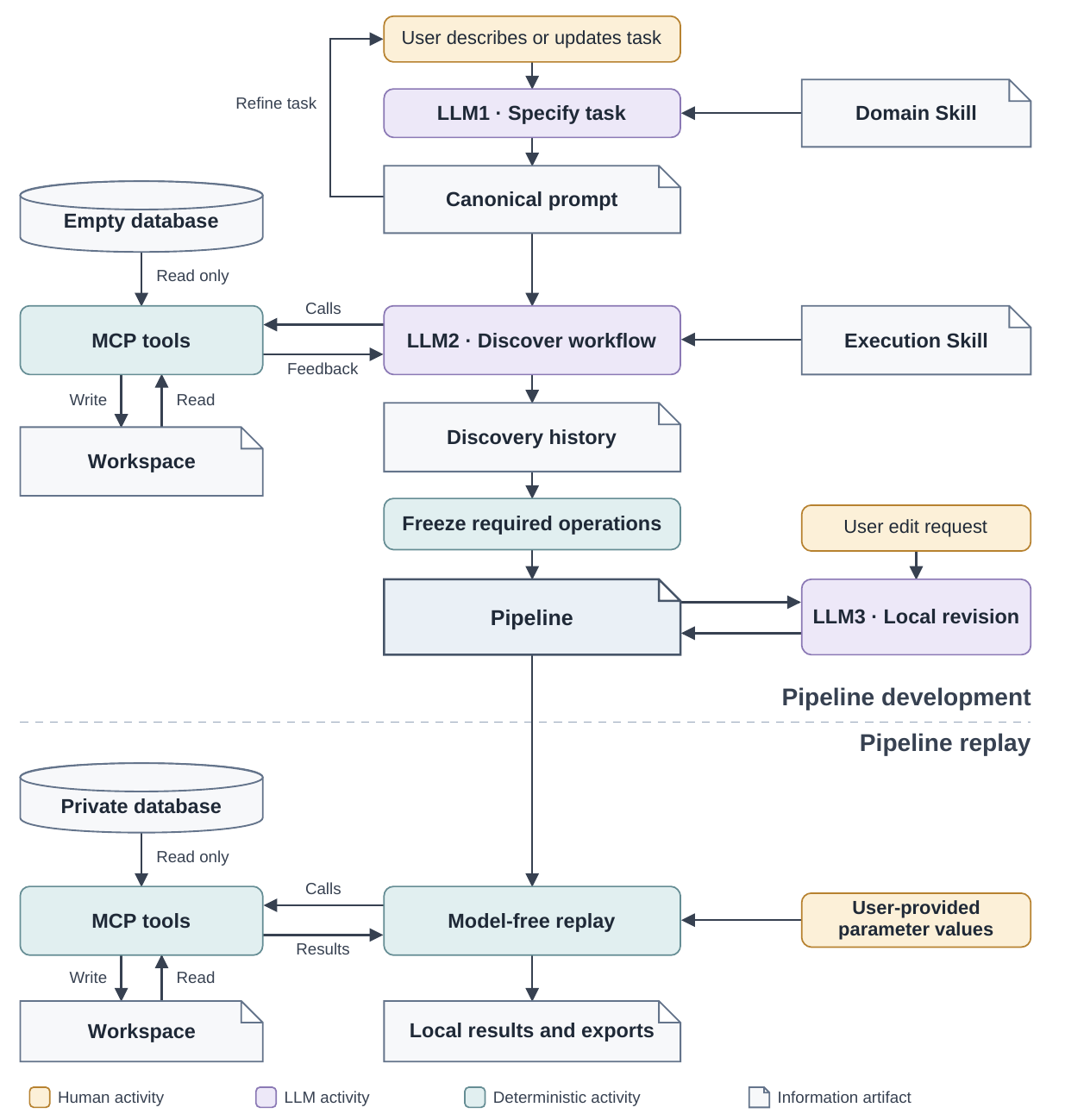}
\caption{Pipeline development (above the dashed line) and routine replay (below). Rounded boxes denote activities, with colour identifying the actor; document and cylinder shapes denote information artifacts and databases. The MCP tools read the database and read and write the workspace in each context. The pipeline includes the canonical task and the executable procedure. Development-time replay and host acceptance of revisions are omitted for clarity.}
\label{fig:architecture}
\end{figure}

Figure~\ref{fig:architecture} separates pipeline development from routine replay. The local host is the software that coordinates model requests, tool calls, and pipelines. LLM1, LLM2, and LLM3 denote roles that may use the same underlying model. Construction uses an empty database and a temporary workspace for intermediate products. Private execution uses the real database and a separate workspace. The tools read their respective databases and can read and write workspace artifacts.

Development also includes private execution of a newly constructed pipeline. The user can inspect its results, refine the task, or repeat discovery. These development-time runs and the host's acceptance checks are omitted from the figure for clarity.

\subsection{The construction environment}
\label{sec:isolation}
The analyst determines which schema definitions, task details, and parameter values may be disclosed to the models. We use \emph{public} for information approved for this purpose; it need not be available to the general public. The host creates the construction database from the approved definitions without copying private rows. Tools invoked during discovery cannot access the real database or its workspace.

MCP provides the interfaces for calling tools and inspecting their descriptions~\cite{mcp2025spec}. The host selects the database and workspace available to each call. This separation gives LLM2 feedback about operations in the construction environment without returning private records or results. The same tool interfaces are used later in the private execution context.

\subsection{From dialogue to a canonical task}
LLM1 turns the user's request into a self-contained \emph{canonical task}: an analytical instruction and named, typed candidate inputs with initial values. The instruction specifies the population, unit of observation, calculation, and requested outputs. A date interval or player identifier can be represented by an input whose value may change on a later run. Candidate inputs become replay parameters only when the retained procedure uses them.

A \emph{Skill} is a set of written instructions supplied to a model role. The domain Skill helps LLM1 interpret terminology and apply conventions relevant to the application. LLM1 specifies the intended analysis; LLM2 chooses the queries and tool sequence. The user can inspect the canonical task, labelled \emph{Canonical prompt} in the figure, and revise it through further dialogue before discovery. The task description remains part of the completed pipeline.

\subsection{Tool discovery on an empty database}
\label{sec:discovery}
LLM2 receives the canonical task, its candidate inputs, and an execution Skill describing how to compose the available operations. It can inspect the approved interfaces and invoke tools one call at a time. An operation can refer symbolically to the result of an earlier operation. The host resolves the reference when making the call, so the model does not need to copy intermediate table contents into tool arguments.

\begin{samepage}
An invalid argument or missing reference can lead LLM2 to correct a call. It can abandon an unsuitable sequence and try another, reusing earlier operations where appropriate. The host records the attempted operations and their outcomes in the discovery history. A call may succeed without contributing to the eventual answer, for example when it prepares a preview that is later replaced.
\par
\end{samepage}

Operations can propose displays or files for delivery to the user. These proposed deliverables are called \emph{staged outputs}; intermediate artifacts used only by other operations are distinct from them. To complete a solution, LLM2 selects the staged outputs that answer the canonical task. Discovery can also end without a usable solution.

A valid query against the construction database will often return no rows. Its result can still carry column names and types needed by subsequent operations. This allows LLM2 to connect operations and check their interfaces. Data-dependent behaviour is exercised when the pipeline runs on populated input.

\subsection{Freezing the selected procedure}
\label{sec:freeze}
The host extracts a pipeline from the discovery history and the selected staged outputs. Each operation record contains its tool name, arguments and symbolic references, completion status, and any staged outputs. Let $R$ be the set of operations producing the selected deliverables. Draw a dependency edge $j\to i$ when an argument of operation $i$ refers to a product of operation $j$. The required operations are
\[
    K=R\ \cup\ \operatorname{ancestors}(R).
\]
The operation producing an intermediate table is retained when a selected output depends on it, even if the table is not itself a deliverable. This selection assumes that all dependencies between operations needed for replay are represented in the recorded references.

The host checks that each required operation succeeded, that its references agree with the recorded dependencies, and that its tool is available to the replay executor. It orders the retained operations so that each operation follows its dependencies, then renumbers the steps and rewrites the references. Only candidate inputs used by the retained operations become replay parameters. Failed attempts and unused operations outside $K$ disappear. The selection preserves the dependencies of the discovered solution; another implementation might use fewer operations.

The pipeline can be represented as
\[
    A=(C,\Theta,P,O),\qquad P=(s_1,\ldots,s_n).
\]
Here $C$ is the canonical task, $\Theta$ declares the used parameters and their defaults, $P$ is the ordered sequence of tool calls, and $O$ identifies the selected outputs and their producing steps. Step arguments consist of literals, parameter references, and references to earlier results. These components can be saved together in a YAML document. The executable hash covers the retained steps, parameter declarations, and output declarations.

\section{Private execution and reuse}
\label{sec:reuse}
\subsection{Executing a fixed sequence}
The host executes the pipeline by binding its parameters and calling the prescribed tools through MCP. References to earlier steps are resolved against fresh results from the current run. For example,
\begin{center}\small
\begin{tabular}{@{}ll@{}}
\playerref & the player supplied for this run \\
\stepref & the artifact produced by step~1 of this run
\end{tabular}
\end{center}
The artifact may have contained an empty table during construction. During private replay it can contain records that another tool consumes without showing them to a model.

Replay makes no model calls. No model selects the next operation, interprets intermediate values, writes the final report, or repairs a private failure. Reporting and export are prescribed tool operations. The host delivers the selected outputs locally after successful execution. If a private call fails, the error is returned locally and is not sent to LLM2 for another attempt.

The orchestration is a finite, straight-line sequence with no pipeline-level conditional branches, loops, recursion, or generation of additional steps at run time. SQL and the tools perform the data-dependent computation. A query can aggregate a variable number of groups, and a tool argument can depend on an earlier result, while the next tool call remains fixed. Section~\ref{sec:case-dataset} shows how a parameter can change the number of exported columns without changing the sequence of calls.

We call a pipeline \emph{accepted} when it has passed structural validation and private replay, including delivery of its selected outputs. The user can inspect the results and save the pipeline. Acceptance records successful execution on those inputs; the analytical method still requires assessment.

\subsection{Repeating an analysis}
\label{sec:three}
An exact replay uses the accepted procedure with the same parameter values. A parameterized replay supplies different values for named parameters for one run. For example, a different date interval changes the selected population while retaining the counting rule, calculation, and report layout. Overrides apply through parameter references and leave the stored defaults unchanged. Neither form of replay calls a model or reconstructs the procedure.

Numerical reproducibility also depends on the data snapshot, tool implementations, ordering where it matters, and relevant computational settings. The pipeline records the prescribed procedure; the data and execution environment determine the results of a particular run.

\subsection{Minor revision and renewed discovery}
Changing a stored default or rearranging a report changes the pipeline itself. Repeating discovery for such a request gives LLM2 latitude to select another query or alter presentation choices elsewhere. During development, incidental changes of this kind motivated a separate editing role.

LLM3 receives the existing pipeline and the edit request, with no tool access or private execution results. It proposes a complete revised pipeline, including the canonical task, or indicates that it cannot make the edit. Its instructions call for the smallest change that satisfies the request. If a default changes, the corresponding canonical input and replay parameter must remain consistent in type and value.

The host validates the candidate's structure and references and replays it privately. Successful execution and delivery of the selected outputs allow the revision to become the current accepted pipeline; failure leaves the previous pipeline available. The user can inspect the changes and results before saving. For a presentation change, comparing numerical outputs and inspecting changes to the calculation helps assess whether the edit preserved the intended analysis.

The user can also revise the canonical task in a completed pipeline and request fresh LLM2 discovery. Repeating discovery with the same task is another way to seek an alternative implementation. These are explicit returns to development, separate from routine replay.

\section{Chess implementation and applications}
\label{sec:chess}
\subsection{Data, tools, and domain instructions}
Chess Agent combines the general harness with a SQLite database of games and positions, analytical tools exposed through MCP, and chess-specific Skills. The harness manages model roles, operation histories, freezing, references, and replay. The domain components provide the schema, analytical operations, and conventions. The examples used OpenAI's \texttt{gpt-6.2-sol} model for the LLM roles.

The database contains game metadata, moves, position features, and stored engine evaluations. Tools support read-only queries, Chess Query Language (CQL) position searches, Stockfish evaluation, calculations on intermediate tables, and presentation or export. The domain Skill helps LLM1 resolve choices such as whose perspective determines a score and whether to count positions or games. The execution Skill guides LLM2 in implementing those choices with the available tools.

Some tools invoke external programs. On empty input, a batch tool can return a typed empty result without starting CQL or Stockfish, allowing subsequent operations to be constructed. Its behaviour on populated input is exercised during private execution. The two applications below use SQL over existing database fields, including stored evaluations; neither retained pipeline invokes CQL or performs a new engine evaluation.

\subsection{Application examples}
\label{sec:cases}
Both applications use the author's game collection. The first produces aggregate outcome statistics; the second exports a dataset for further analysis. The player identifier is shown as \texttt{my\_name} in the prompts and pipeline listings. The appendices reproduce the canonical tasks, parameters, retained steps, and output declarations. For the second application, Appendix~\ref{app:dataset_generation} also includes the discovery trace, showing how the pipeline was constructed and which operations were eliminated during freezing.

\subsubsection{Outcome statistics for pawn endings}
\label{sec:case-pawn}
The first application asks for results conditional on reaching a pawn ending. The input to LLM1 was:

\begin{TaskPrompt}
Find first occurrence of pawn endings in my_name's games. give the outcome statistics for these games. stratify the games by evaluation of the first qualifying position, labelled as worse (eval <= -1 pawn unit), equal (eval between -1 and +1 pawn) and better (eval >= 1 pawn unit). Show the statistics as a table with one row per stratum, on the screen.
\end{TaskPrompt}

LLM1 expanded the request into a canonical task that specifies which position to select, how to orient evaluations and outcomes, and how to handle missing values. Appendix~\ref{app:pawn_endings} contains this task and the three retained calls: resolve the player identifier, query the database into a workspace table, and display that table. The query performs the selection and statistical calculations; no separate chess-search or engine call is needed.

For each game, the query selects the smallest ply at which no queens, rooks, bishops, or knights of either colour remain. Pawn counts are unrestricted. A game contributes at most once, however long the ending lasts. The stored evaluation at that position is converted from White's perspective to the selected player's perspective, reversing its sign when the player is Black. Games lacking a centipawn evaluation at the first qualifying position are excluded rather than represented by a later position. Unrecognized or missing game results are also excluded from the outcome statistics.

With evaluation $e$ in pawn units, the default strata are $e\leq-1$ (worse), $-1<e<1$ (equal), and $e\geq1$ (better). This makes the boundary values unambiguous. Wins, draws, and losses are scored as $1,\tfrac12,0$ from the selected player's perspective. For $n$ games in a stratum, the pipeline reports the mean score $\bar Y$ and the conventional standard error $s/\sqrt n$, using the sample standard deviation $s$. These standard errors do not adjust for dependence between games.

\begin{table}[!htbp]
\centering\small
\begin{tabular}{@{}lrrrrr@{}}
\toprule
Stratum & Wins & Draws & Losses & Mean game score & Standard error \\
\midrule
worse  &   7 &   9 & 86 & 0.113 & 0.02779 \\
equal  &  56 & 335 & 28 & 0.533 & 0.01083 \\
better & 215 &   5 &  4 & 0.971 & 0.01007 \\
\bottomrule
\end{tabular}
\caption{Reported outcomes stratified by the stored evaluation of the first qualifying pawn-ending position, using the pipeline's default parameters. Each game contributes once.}
\label{tab:pawn_endings}
\end{table}

The run contains 745 games across the three strata (Table~\ref{tab:pawn_endings}). Mean score rises from 0.113 in the worse stratum to 0.971 in the better stratum. These are descriptive results conditional on the selected position and available evaluation. The strata do not supply calibrated expected game scores, so the table alone does not establish overperformance relative to the engine assessment.

Freezing retains all three calls: the displayed result depends on the query, which depends on player resolution. The player identifier and evaluation thresholds remain parameters, so their values can change on replay without changing the procedure.


\subsubsection{Preparing a dataset for further analysis}
\label{sec:case-dataset}

The second application prepares a local dataset for subsequent statistical or
machine-learning work. The request to LLM1 was:

\begin{TaskPrompt}
Find the blitz games where my_name plays black that are at least N=10 ply long, played between 1. jan 2024 and 31. dec 2025. Create a dataset with one row for each game and rows in the following order: outcome (1, 0.5 or 0), Time, one row for each of the N firs plies in SAN notation, stored eval of the position after N ply, opponent name, opponent rating. Sort the rows with oldest first. Write the table to research_table.csv. On the screen report the number of rows and display the top 5 rows. Anonymize the opponents names to opponent_001, opponent_002 etc.
\end{TaskPrompt}

The canonical task resolves the requested fields into columns and specifies
player perspective, missing values, and chronological ordering. With the default
parameters, it selects the author's blitz games as Black from 1 January 2024
through 31 December 2025, with at least $N=10$ plies (half-moves). Each row contains
the outcome, recorded date and available time, the first $N$ plies in standard
algebraic notation (SAN), the stored evaluation immediately after ply~$N$, and
the opponent's alias and rating. Outcomes and evaluations use the selected
player's perspective. Missing outcomes, evaluations, and ratings are left blank.
No new engine evaluation is requested.

Opponent names are replaced by pseudonyms assigned in order of first appearance
in the sorted dataset. The same opponent receives the same alias within a run;
the aliases are recomputed for each selected population. Game identifiers break
ties between identical timestamps. Neither the original opponent names nor the
alias mapping is included in the delivered outputs.

\paragraph{Discovery and freezing.}
LLM2 attempted 14 tool calls against the empty construction database and its
workspace. Two attempts to construct a query with $N$ move columns failed because
the workspace SQL tool rejected their common-table-expression names. LLM2 then
used the database SQL tool to construct the query text from $N$ and a fixed
template. This computation requires no game records. A later preview attempt
failed because it referred to a \texttt{rowid} column absent from the workspace
table. LLM2 recovered by constructing a preview query from the successful dataset
query.

The discovery history contains three failed calls and one successful but unused
call. The latter constructed a preview query with a literal row limit; the chosen
alternative used a bound parameter. Freezing removes these four calls and retains
the ten operations required for the CSV file, row count, and preview. Thus both
unsuccessful attempts and a successful alternative disappear from the executable
procedure. Appendix~\ref{app:dataset_generation} reproduces the discovery
trace and the final pipeline, with a mapping between their step identifiers.

\paragraph{Private execution and minor revision.}
The initial private run produced 3,064 rows and 15 columns in
\texttt{research\_table.csv}, together with a row count and a five-row preview. The displayed SAN
values contained unwanted surrounding quotation marks, for example
\verb|"Nc3"|. The author therefore asked LLM3:

\begin{TaskPrompt}
display the moves without surrounding "s.
\end{TaskPrompt}

The final pipeline contains one localized executable change: a \texttt{trim}
expression removes surrounding double quotes from each extracted SAN value.
The canonical task is updated accordingly. Comparing the final artifact with
the retained discovery calls, after accounting for renamed input references and
renumbered steps, shows that the other nine calls are unchanged. Both the full
dataset and its preview use the revised query construction. Private execution
after the edit again reported 3,064 rows; the other values in the five displayed
rows were unchanged. Table~\ref{tab:ml_dataset} shows this revised preview.

\begin{table}[!htbp]
\centering\small
\setlength{\tabcolsep}{4pt}
\begin{tabular}{@{}rrlrlr@{}}
\toprule
Row & Outcome & Time & Eval.\ (cp) & Opponent & Rating \\
\midrule
1 & 1.0 & 2024-01-01 13:22:48 & -89 & opponent\_001 & 2340 \\
2 & 0.5 & 2024-01-01 13:32:32 & -35 & opponent\_002 & 2417 \\
3 & 0.5 & 2024-01-01 15:27:27 & -45 & opponent\_003 & 2473 \\
4 & 1.0 & 2024-01-01 15:55:39 & 36 & opponent\_004 & 2443 \\
5 & 1.0 & 2024-01-01 18:28:26 & -55 & opponent\_005 & 2357 \\
\bottomrule
\end{tabular}

\vspace{8pt}
\begin{tabular}{@{}r*{10}{c}@{}}
\toprule
Row & Ply\_1 & Ply\_2 & Ply\_3 & Ply\_4 & Ply\_5 & Ply\_6 & Ply\_7 & Ply\_8 & Ply\_9 & Ply\_10 \\
\midrule
1 & Nc3 & Nf6 & e4 & d6 & d4 & g6 & f4 & Bg7 & Nf3 & c5 \\
2 & c4 & e6 & e4 & d5 & cxd5 & exd5 & exd5 & Qxd5 & Nc3 & Qd8 \\
3 & d4 & Nf6 & Nf3 & g6 & c4 & Bg7 & g3 & O-O & Bg2 & d6 \\
4 & Nf3 & d5 & g3 & c5 & Bg2 & Nc6 & O-O & e5 & d3 & Be7 \\
5 & Nf3 & Nf6 & c4 & g6 & Nc3 & Bg7 & e4 & d6 & d4 & O-O \\
\bottomrule
\end{tabular}
\caption{The five-row preview after the LLM3 revision, split into two panels.
The row index is added for presentation. In the CSV, the move columns follow
\texttt{Time}, followed by \texttt{Evaluation\_cp}, \texttt{Opponent}, and
\texttt{Opponent\_rating}.}
\label{tab:ml_dataset}
\end{table}

\paragraph{Parameterized replay.}
The revised pipeline was subsequently run with
\texttt{start\_date="1.\ jan 2025"} and \texttt{ply\_count=4}, leaving the other
parameters unchanged. This run produced 1,158 rows and nine columns
(Table~\ref{tab:dataset-runs}). The stored defaults remained at the original
2024--2025 period and $N=10$; the new values applied to this run only.

\begin{table}[!htbp]
\centering\small
\begin{tabular}{@{}llrrr@{}}
\toprule
Run & Period & $N$ & Rows & Columns \\
\midrule
Initial pipeline & 2024--2025 & 10 & 3,064 & 15 \\
After LLM3 revision & 2024--2025 & 10 & 3,064 & 15 \\
Parameterized replay & 2025 & 4 & 1,158 & 9 \\
\bottomrule
\end{tabular}
\caption{Observed outputs during development and reuse. Periods cover complete
calendar years. The final run changes both the start date and the ply count;
its end date remains 31 December 2025.}
\label{tab:dataset-runs}
\end{table}

The parameter $N$ determines the minimum game length, the number of move columns,
and the position from which the stored evaluation is taken. Changing it therefore
changes the output schema as well as part of the selection rule. SQL performs
this computation within the same ten-call pipeline; parameterized replay requires
no further model invocation. The exported dataset can then be used for feature
encoding, model fitting, or other analysis in separate local software. These
subsequent methods are outside the example. All populated tables, counts, and
exports are produced locally without being passed to the models by the execution
path.

\FloatBarrier

\section{Related work}
\label{sec:related}
The closest precedents combine language-model construction with restricted data access or a reusable executable artifact. Scientific workflow systems provide the longer history of recording how analyses are constructed and revised.

\paragraph{Constructing analyses without exposing records.}
FlowMind generates financial workflows from domain and API descriptions while keeping proprietary records and implementations outside direct LLM access. Users can inspect the workflow and request revisions~\cite{zeng2023flowmind}. Data-Copilot separates intent analysis from interface composition and connects data acquisition, processing, and visualization through explicit dependencies~\cite{zhang2024datacopilot}. Metadata-Driven AI generates a reusable program from metadata and executes it beside the data, avoiding a new model invocation on each run~\cite{pawlak2026metadata}. These are close precedents for the motivation and separation of construction from execution.

In their SQL-agent case study, Beurer-Kellner et al.\ describe generated code processing database contents without passing those contents through an LLM~\cite{beurerkellner2025patterns}. The TRACE technical proposal similarly separates model-assisted development using approved descriptions and sanitized feedback from stand-in data from execution on protected records~\cite{mueller2026trace}. CaMeL and \emph{tacit} address related security questions through interpreted programs and capabilities that govern information flows and effects~\cite{debenedetti2026camel,odersky2026tacit}. In our harness, the host and trusted tools enforce the construction and execution boundary.

\paragraph{Retaining executable procedures.}
VisTrails records changes to scientific workflows, Galaxy can extract workflows from exploratory histories, and Snakemake makes computational procedures explicit and reusable~\cite{freire2006vistrails,goecks2010galaxy,molder2021snakemake}. Dunn et al.\ discuss reproducibility in model-assisted science in terms of whether models remain on the data path when an analysis is rerun~\cite{dunn2026reproducible}. AgentRR explores recording and replaying agent experience, including separation of untrusted exploration from trusted execution~\cite{feng2025agentrr}. TraceForge reconstructs workflow graphs from observed runs and uses a human gate to freeze a version~\cite{rajan2026traceforge}.

Compiled AI produces executable artifacts during a model-assisted compilation phase and runs them without further model invocation~\cite{trooskens2026compiledai}. ReTrace distils tool-use traces into planner-free programs~\cite{hou2026retrace}. These systems differ in how they obtain the retained program and whether execution can still invoke models. Our harness exercises tools against an empty database, retains the dependencies of selected outputs, and requires model-free replay through reporting, export, and failure handling.

Scientific tool composition through MCP also has precedents. MPDAS translates tool chains into Snakemake workflows for single-cell analyses and uses models to interpret outputs~\cite{hayakawa2026mpdas}. Paper2Agent builds and tests MCP tools from scientific papers and their code so that agents can apply the methods~\cite{miao2026paper2agent}. Here MCP supplies the interface for both construction and replay; the host determines the data available to each call.

\paragraph{Revising an established analysis.}
Flowco supports node-scoped synthesis, dependency-aware updates, and locks on nodes that a model cannot change~\cite{freund2025flowco}. Data Interpreter preserves successful upstream work when revising a plan, and Data Formulator~2 represents branching histories of visual analyses~\cite{hong2025datainterpreter,wang2025dataformulator2}. TabClean synthesizes reusable cleaning programs and supports incremental repair and rollback~\cite{wang2026tabcleanworkshop}. Our editing role supports minor requests such as changing a default or rearranging a report. It receives the existing pipeline, has no tools, and is instructed to minimize the change. Flowco's locks offer a stronger mechanical constraint on which components may be altered.

\section{Discussion}
\label{sec:discussion}
\subsection{Data isolation and its scope}
The harness keeps private records, intermediate products, analytical results, and execution errors outside model processing. Its construction environment contains the approved structure without private rows, and its private execution path makes no model calls. This removes the need to provide a redacted or anonymized copy of the records to an external model service.

The analyst determines what may be disclosed in schema definitions, task descriptions, parameter values, and subsequent requests. Information conveyed through those inputs is outside the confidentiality objective of the private execution path. The host, tools, and local environment are trusted to enforce the separation and control tool effects, including output destinations.

This boundary does not establish formal noninterference across the
whole interactive lifecycle~\cite{sabelfeld2003ifc}.
The user has access to private data and may therefore disclose
information to the LLMs.
This may happen explicitly, if the user, possibly inadvertently,
includes private information in a prompt.
It may also happen indirectly, if the user defines a follow-up
pipeline or modifies an existing one in response to the results
of a previous analysis.
Such interactions could in principle enable the external model
provider to infer private information.

\subsection{Robust execution and analytical scope}
A fixed sequence makes the orchestration available for inspection before private execution and avoids further model decisions during a run. Reuse preserves the chosen procedure, and revisions are explicit changes to a stored artifact. Conditional computation and iteration occur inside SQL and trusted tools, whose behaviour can be tested independently.

The dataset application illustrates this distinction: SQL constructs the requested move columns within a fixed sequence of ten calls. The parameterized replay demonstrates that both the population and the structure of the resulting dataset can change without altering this sequence. The restriction concerns orchestration, not the size of the data or the computation within a tool.

The model constructs a procedure; the analyst interprets its private results and decides what to investigate next. A local export also allows the work to continue in software whose methods are not provided by the harness. The dataset example prepares inputs for that work without attempting to fit or validate a machine-learning model.

An empty construction database cannot expose every analytical error. A join that returns an empty, well-typed table can multiply observations on populated data. Small public test cases with known answers and checks against independent calculations can assess these behaviours. For stored engine evaluations, reproducing and interpreting results also requires knowledge of how the evaluations were produced, including relevant engine versions and search settings.

\subsection{What the applications establish}
Chess provides a comparatively low-risk setting for studying a harness motivated by sensitive-data applications. Its analytical complexity remains substantial: an analysis must connect games to positions, define the observation counted, orient scores to a player, and handle missing values. The examples make these choices visible in the canonical tasks and retained queries.

The pawn-ending pipeline produces a report from three calls. In the dataset application, freezing reduces a discovery history of 14 calls to a pipeline of ten calls, eliminating three failed attempts and one successful but unused operation. A minor revision through LLM3 corrects the formatting of the moves without changing the other retained operations. Subsequent replay with different parameters produces a dataset with a different number of rows and columns, without further model involvement. These examples illustrate construction, freezing, revision, and replay; they do not estimate success rates or editing reliability across a task population, or performance at larger scales.

A direct comparison would give another model the same approved schema, tool descriptions, task, and permitted feedback and ask it to write a reusable script for local execution. Such a script also allows model-free reruns. Relevant outcomes include successful execution, independently checked analytical results, the effort needed to obtain the procedure, and preservation of quantities that an edit should leave unchanged. Repeated construction and editing trials would be needed for that comparison.

The current evidence comes from one domain. Applying the harness elsewhere requires suitable schemas, tools, Skills, and validation cases. The chess examples examine construction and reuse without requiring sensitive institutional records. They do not by themselves validate deployment under another domain's privacy or operational requirements.

\section{Conclusion}
The harness uses externally hosted models to construct analyses while keeping private execution local and model-free. Exploration against an empty database produces an operation history; freezing retains the dependencies of selected outputs as a reusable pipeline. The canonical task, parameters, and prescribed calls remain available for inspection, replay, and minor revision. The chess applications demonstrate both an aggregate analysis and preparation of a research dataset. They provide concrete examples of a general construction mechanism whose domain-specific requirements are supplied through schemas, tools, and instructions.

\section*{Software availability}
A public GitHub release of the general harness and the chess-specific components is planned.

\section*{Declaration of AI assistance}
Under the author's direction, ChatGPT and Codex assisted with implementation, debugging, testing, and software documentation. They also assisted with literature research, drafting, language revision, and manuscript preparation. The author is responsible for the implementation claims, citations, methods, results, and conclusions.

\bibliographystyle{unsrtnat}
\begingroup
\footnotesize
\setlength{\bibsep}{3.5pt}
\interlinepenalty=10000
\bibliography{references}
\endgroup

\clearpage
\appendix
\section{Pawn endings pipeline}
\label{app:pawn_endings}

\begin{Verbatim}[breaklines=true,
                 breakanywhere=true,
                 breaksymbolleft={},
                 fontsize=\footnotesize]
canonical_task: 'Using the chess research database, select games played by ${inputs.player_identifier}
  as either White or Black. For each game, find the earliest position by ply with
  both kings present, no queens, rooks, bishops, or knights, and no restriction on
  pawn counts. Retain one observation per qualifying game, regardless of how many
  qualifying positions it contains.
  Use the stored evaluation of that first qualifying position from the selected player''s
  perspective, not necessarily the side to move. Use an already player-relative evaluation
  as recorded; if only a documented White-relative evaluation is available, reverse
  its sign exactly once when the selected player is Black. Do not change the position
  itself. In pawn units (100 centipawns per pawn), assign the game to worse if the
  evaluation is less than or equal to ${inputs.worse_eval_threshold_pawns}, equal
  if it is strictly between ${inputs.worse_eval_threshold_pawns} and ${inputs.better_eval_threshold_pawns},
  or better if it is greater than or equal to ${inputs.better_eval_threshold_pawns}.
  If the first qualifying position has no usable evaluation, omit the game from the
  strata rather than substituting a later position.
  For each stratum, calculate wins, draws, losses, mean game score, and the standard
  error of the mean from the selected player''s perspective. Use recognized normalized
  results when available; otherwise interpret 1-0 as a White win, 0-1 as a Black win,
  and 1/2-1/2 as a draw. Exclude missing or unrecognized results from all outcome
  calculations. Score wins as 1, draws as 0.5, and losses as 0. Calculate the arithmetic
  mean and use the sample standard deviation of game scores divided by the square
  root of the number of recognized results for the standard error. Show undefined
  statistics as unavailable.
  Display only the statistics table on screen, with one row for each stratum in this
  order: worse, equal, better. Include the stratum label, win count, draw count, loss
  count, mean game score to 3 decimal places, and standard error to 5 decimal places.
  Round only for display.'
  
parameters:
  player_identifier:
    type: string
    default: my_name
  worse_eval_threshold_pawns:
    type: float
    default: -1.0
  better_eval_threshold_pawns:
    type: float
    default: 1.0
    
steps:
- id: step_1
  tool: resolve_player_identifier
  arguments:
    player_identifier: ${parameters.player_identifier}
- id: step_2
  tool: query_database_to_workspace
  arguments:
    sql: WITH first_match AS (SELECT p.game_id, MIN(p.ply) AS ply FROM games AS g
      JOIN positions AS p ON p.game_id = g.game_id WHERE (g.white_player = :participant
      OR g.black_player = :participant) AND p.white_queen_count = 0 AND p.black_queen_count
      = 0 AND p.white_rook_count = 0 AND p.black_rook_count = 0 AND p.white_bishop_count
      = 0 AND p.black_bishop_count = 0 AND p.white_knight_count = 0 AND p.black_knight_count
      = 0 GROUP BY p.game_id), oriented AS (SELECT (CASE WHEN g.white_player = :participant
      THEN p.eval_cp_white ELSE -p.eval_cp_white END) / 100.0 AS eval_pawns, CASE
      WHEN g.result = '1/2-1/2' THEN 0.5 WHEN g.result = '1-0' THEN CASE WHEN g.white_player
      = :participant THEN 1.0 ELSE 0.0 END WHEN g.result = '0-1' THEN CASE WHEN g.white_player
      = :participant THEN 0.0 ELSE 1.0 END END AS score FROM first_match AS m JOIN
      games AS g ON g.game_id = m.game_id JOIN positions AS p ON p.game_id = m.game_id
      AND p.ply = m.ply WHERE p.eval_cp_white IS NOT NULL), bucketed AS (SELECT CASE
      WHEN eval_pawns <= :worse THEN 1 WHEN eval_pawns >= :better THEN 3 WHEN eval_pawns
      > :worse AND eval_pawns < :better THEN 2 END AS bucket, score FROM oriented
      WHERE score IS NOT NULL), stats AS (SELECT bucket, COUNT(CASE WHEN score = 1.0
      THEN 1 END) AS wins, COUNT(CASE WHEN score = 0.5 THEN 1 END) AS draws, COUNT(CASE
      WHEN score = 0.0 THEN 1 END) AS losses, COUNT(*) AS n, AVG(score) AS mean_score,
      SUM(score * score) AS sum_squares, SUM(score) AS score_sum FROM bucketed GROUP
      BY bucket), strata(bucket, stratum) AS (VALUES (1, 'worse'), (2, 'equal'), (3,
      'better')) SELECT s.stratum, COALESCE(a.wins, 0) AS wins, COALESCE(a.draws,
      0) AS draws, COALESCE(a.losses, 0) AS losses, CASE WHEN a.n > 0 THEN printf('%.3f',
      a.mean_score) ELSE 'Unavailable' END AS mean_game_score, CASE WHEN a.n > 1 THEN
      printf('%.5f', sqrt(max(0.0, a.sum_squares - a.score_sum * a.score_sum / a.n)
      / (1.0 * a.n * (a.n - 1)))) ELSE 'Unavailable' END AS standard_error FROM strata
      AS s LEFT JOIN stats AS a ON a.bucket = s.bucket ORDER BY s.bucket
    parameters:
      participant: ${steps.step_1.scalar_value}
      worse: ${parameters.worse_eval_threshold_pawns}
      better: ${parameters.better_eval_threshold_pawns}
    summary: Outcome statistics by first qualifying position evaluation stratum
- id: step_3
  tool: present_table
  arguments:
    source_artifact: ${steps.step_2.artifact_id}
    columns:
    - stratum
    - wins
    - draws
    - losses
    - mean_game_score
    - standard_error
    column_labels:
      stratum: Stratum
      wins: Wins
      draws: Draws
      losses: Losses
      mean_game_score: Mean game score
      standard_error: Standard error
outputs:
- step_id: step_3
  effect_index: 0
  effect_type: screen
\end{Verbatim}
\clearpage

\section{Dataset development trace and final pipeline}
\label{app:dataset_generation}

This appendix records the development of the dataset application in
Section~\ref{sec:case-dataset}. The discovery trace precedes the LLM3 revision;
the final pipeline includes that revision. The player identifier is displayed as
\texttt{my\_name} throughout, as in the main text. Run identifiers, artifact
identifiers, and hashes in the trace refer to the original record.

\subsection{Discovery calls and retained operations}

Table~\ref{tab:dataset-steps} maps the original discovery calls to the ten steps
in the final pipeline. The CSV, count, and preview are produced by discovery
calls~8, 9, and~14. Their dependency closure is
\[
K=\{1,2,3,6,7,8,9,12,13,14\}.
\]
Calls~4, 5, and~10 failed. Call~11 succeeded, but no selected output depends on
it. These four calls are absent from the final pipeline.

\begin{table}[!htbp]
\centering\small
\setlength{\tabcolsep}{4pt}
\begin{tabularx}{\linewidth}{@{}rrYll@{}}
\toprule
Call & Step & Operation & Uses calls & Disposition \\
\midrule
1 & 1 & Resolve player identifier & -- & Retained \\
2 & 3 & Extract eligible games and moves & 1 & Retained \\
3 & 5 & Orient rows to the player & 1, 2 & Retained \\
4 & -- & Construct SQL with a recursive expression & 3 & Failed \\
5 & -- & Construct SQL with an alternative expression & 3 & Failed \\
6 & 2 & Construct SQL for $N$ move columns & -- & Retained \\
7 & 6 & Produce the dataset table & 3, 6 & Retained \\
8 & 8 & Export the dataset & 7 & Retained \\
9 & 9 & Display the dataset row count & 7 & Retained \\
10 & -- & Select a preview using \texttt{rowid} & 7 & Failed \\
11 & -- & Construct a preview query with a literal limit & 6 & Unused \\
12 & 4 & Construct a preview query with a bound limit & 6 & Retained \\
13 & 7 & Produce the preview table & 3, 12 & Retained \\
14 & 10 & Display the preview & 13 & Retained \\
\bottomrule
\end{tabularx}
\caption{Discovery calls, their recorded references to earlier calls, and their
corresponding steps in the final pipeline. Parameter references are omitted from
the dependency column. The final numbering follows an order consistent with the
dependencies.}
\label{tab:dataset-steps}
\end{table}

The count uses the row-count metadata of the dataset artifact. The preview
executes a limited version of the dataset query against the same oriented source
table. It therefore shares the query construction and ordering rule with the
full export.

\subsection{Discovery trace}

The following listing reproduces the discovery trace, including
failed calls and the successful unused alternative. Some tool responses were
abbreviated in the recorded trace and retain their original
\texttt{[truncated for display]} markers. This is a tool-call trace, rather than a
complete transcript of model inputs and outputs. The zero-row results belong to
construction against the empty database; the scalar query strings are generated
from parameters and fixed templates. The populated previews and counts reported
in the main text come from separate private executions.

\begin{Verbatim}
Run: analysis_08ca93ef4cfc46e384e3c916f2ad0cac
Canonical source: compiled canonical (compilation compilation_40c617f4210d4a88bab7e87f7cfc207b) [hash 361570055b32f2350c3da514d2d54b93c0d7525cdf8877a6067f59d3539bba8d]
Canonical prompt: Use the chess research database to select blitz games in which ${inputs.player_identifier} played as ${inputs.player_color}. Include games played from ${inputs.start_date} through ${inputs.end_date}, inclusive, with at least ${inputs.ply_count} plies. Produce one row per qualifying game, sorted by game timestamp from oldest to newest, with a deterministic tie-breaker for identical timestamps.

Use these columns in order: (1) outcome from the selected player's perspective: win = 1, draw = 0.5, loss = 0, and missing or unrecognized results left blank; (2) Time, the recorded game timestamp; (3) one column per ply for the first ${inputs.ply_count} plies, in play order and in SAN; (4) the stored evaluation of the position immediately after ply ${inputs.ply_count}, expressed from the selected player's perspective; (5) anonymized opponent name; (6) the opponent's recorded rating for that game. For the evaluation, reverse the sign of a documented White-relative value only when the selected player is Black, and do not negate a value already expressed from the selected player's perspective. Use the stored evaluation as recorded without adding engine-run or depth restrictions. Leave missing evaluations and ratings blank rather than imputing them.

Assign opponent aliases sequentially in order of first appearance in the sorted dataset: opponent_001, opponent_002, and so on. Reuse the same alias for the same opponent across games. Do not expose original opponent names or the alias mapping in either output.

Write the complete sorted dataset to ${inputs.output_filename}. Separately, on screen, report the total number of game rows, excluding the header, and display only the first ${inputs.preview_row_count} rows of the same sorted dataset, or all rows if fewer qualify. Do not add other user-facing outputs.
Canonical inputs: {"end_date": {"type": "string", "value": "31. dec 2025"}, "output_filename": {"type": "string", "value": "research_table.csv"}, "player_color": {"type": "string", "value": "black"}, "player_identifier": {"type": "string", "value": "my_name"}, "ply_count": {"type": "integer", "value": 10}, "preview_row_count": {"type": "integer", "value": 5}, "start_date": {"type": "string", "value": "1. jan 2024"}, "time_control": {"type": "string", "value": "blitz"}}

Canonical source matches execution_start: yes

Tools: analyze_fen, analyze_positions_stockfish, apply_moves_to_positions, create_chess_game_report, create_chess_position_report, create_game_pgn, create_html_table_report, export_workspace_dataset, filter_positions_cql, moves_to_position_key, orient_game_rows, present_scalar, present_table, preview_workspace_artifact, query_database_readonly, query_database_to_workspace, query_workspace_readonly, resolve_player_identifier, select_games_by_opening_prefix, summarize_weighted_distributions, summarize_workspace_table
Resources: schema://database, schema://games, schema://positions, schema://move_semantics, schema://recovery_events, cql://research-v1, player-perspective://readonly, workspace-sql://readonly, workspace-stockfish://batch-analysis, workspace-html://table-report, workspace-chess://position-report, workspace-chess://game-report
Skills: chess-research-conventions (1b0f1a611470)
Iteration 1 (? ms): V2 deterministic tool selection
Tool step_1: resolve_player_identifier
  arguments: {"player_identifier": "${inputs.player_identifier}"}
  resolved_arguments: {"player_identifier": "my_name"}
  result: {"is_complete": true, "is_truncated": false, "result_type": "scalar", "scalar_value": "my_name", "status": "ok"}
  artifacts: {"artifact_metadata": {}, "artifact_references": {}}
Iteration 2 (? ms): V2 deterministic tool selection
Tool step_2: query_database_to_workspace
  arguments: {"parameters": {"end_date": "${inputs.end_date}", "player": "${steps.step_1.scalar_value}", "ply_count": "${inputs.ply_count}", "side": "${inputs.player_color}", "speed": "${inputs.time_control}", "start_date": "${inputs.start_date}"}, "sql": "WITH date_inputs(kind, raw_date) AS (VALUES ('start', :start_date), ('end', :end_date)), date_parts AS (SELECT kind, trim(raw_date) AS raw_date, instr(trim(raw_date), '.') AS dot_pos FROM date_inputs), date_limits AS (SELECT kind, coalesce(date(replace(raw_date, '.', '-')), date(printf('%04d-%02d-%02d', cast(substr(raw_date, -4) AS integer), CASE lower(substr(trim(substr(raw_date, dot_pos + 1)), 1, 3)) WHEN 'jan' THEN 1 WHEN 'feb' THEN 2 WHEN 'mar' THEN 3 WHEN 'apr' THEN 4 WHEN 'may' THEN 5 WHEN 'jun' THEN 6 WHEN 'jul' THEN 7 WHEN 'aug' THEN 8 WHEN 'sep' THEN 9 WHEN 'oct' THEN 10 WHEN 'nov' THEN 11 WHEN 'dec' THEN 12 END, cast(substr(raw_date, 1, dot_pos - 1) AS integer)))) AS iso_day FROM date_parts), candidate_games AS (SELECT g.game_id, g.white_player, g.black_player, g.result, g.white_elo, g.black_elo, coalesce(date(replace(g.utc_date, '.', '-')), date(replace(g.date, '.', '-'))) AS game_day, CASE WHEN time(g.utc_time) IS NOT NULL THEN g.utc_time END AS game_time FROM games g WHERE g.time_control_category = :speed AND ((:side = 'white' AND g.white_player = :player) OR (:side = 'black' AND g.black_player = :player))) SELECT g.game_id, g.white_player, g.black_player, g.result, g.white_elo, g.black_elo, CASE WHEN g.game_time IS NULL THEN g.game_day ELSE g.game_day || ' ' || g.game_time END AS game_timestamp, pos.eval_cp_white, (SELECT json_group_object(m.ply, m.san_move_played) FROM positions m WHERE m.game_id = g.game_id AND m.ply BETWEEN 1 AND :ply_count) AS san_by_ply FROM candidate_games g JOIN positions pos ON pos.game_id = g.game_id AND pos.ply = :ply_count WHERE g.game_day BETWEEN (SELECT iso_day FROM date_limits WHERE kind = 'start') AND (SELECT iso_day FROM date_limits WHERE kind = 'end')", "summary": "Eligible games with stored opening SAN and post-ply evaluation"}
  resolved_arguments: {"parameters": {"end_date": "31. dec 2025", "player": "my_name", "ply_count": 10, "side": "black", "speed": "blitz", "start_date": "1. jan 2024"}, "sql": "WITH date_inputs(kind, raw_date) AS (VALUES ('start', :start_date), ('end', :end_date)), date_parts AS (SELECT kind, trim(raw_date) AS raw_date, instr(trim(raw_date), '.') AS dot_pos FROM date_inputs), date_limits AS (SELECT kind, coalesce(date(replace(raw_date, '.', '-')), date(printf('%04d-%02d-%02d', cast(substr(raw_date, -4) AS integer), CASE lower(substr(trim(substr(raw_date, dot_pos + 1)), 1, 3)) WHEN 'jan' THEN 1 WHEN 'feb' THEN 2 WHEN 'mar' THEN 3 WHEN 'apr' THEN 4 WHEN 'may' THEN 5 WHEN 'jun' THEN 6 WHEN 'jul' THEN 7 WHEN 'aug' THEN 8 WHEN 'sep' THEN 9 WHEN 'oct' THEN 10 WHEN 'nov' THEN 11 WHEN 'dec' THEN 12 END, cast(substr(raw_date, 1, dot_pos - 1) AS integer)))) AS iso_day FROM date_parts), candidate_games AS (SELECT g.game_id, g.white_player, g.black_player, g.result, g.white_elo, g.black_elo, coalesce(date(replace(g.utc_date, '.', '-')), date(replace(g.date, '.', '-'))) AS game_day, CASE WHEN time(g.utc_time) IS NOT NULL THEN g.utc_time END AS game_time FROM games g WHERE g.time_control_category = :speed AND ((:side = 'white' AND g.white_player = :player) OR (:side = 'black' AND g.black_player = :player))) SELECT g.game_id, g.white_player, g.black_player, g.result, g.white_elo, g.black_elo, CASE WHEN g.game_time IS NULL THEN g.game_day ELSE g.game_day || ' ' || g.game_time END AS game_timestamp, pos.eval_cp_white, (SELECT json_group_object(m.ply, m.san_move_played) FROM positions m WHERE m.game_id = g.game_id AND m.ply BETWEEN 1 AND :ply_count) AS san_by_ply FROM candidate_games g JOIN positions pos ON pos.game_id = g.game_id AND pos.ply = :ply_count WHERE g.game_day BETWEEN (SELECT iso_day FROM date_limits WHERE kind = 'start') AND (SELECT iso_day FROM date_limits WHERE kind = 'end')", "summary": "Eligible games with stored opening SAN and post-ply evaluation"}
  result: {"artifact_complete": true, "artifact_id": "ws_20261128_c44855cfb6f1", "artifact_status": "active", "column_names": ["game_id", "white_player", "black_player", "result", "white_elo", "black_elo", "game_timestamp", "eval_cp_white", "san_by_ply"], "columns": [{"name": "game_id", "type": "null"}, {"name": "white_player", "type": "null"}, {"name": "black_player", "type": "null"}, {"name": "result", "type": "null"}, {"name": "white_elo", "type": "null"}, {"name": "black_elo", "type": "null"}, {"name": "game_timestamp", "type": "null"}, {"name": "eval_cp_white", "type": "null"}, {"name": "san_by_ply", "type": "null"}], "complete": true, "content_hash": "5fadb367fd4edaa02140ada5845148cc6a8a3a05a619a980b85de1ec39746d1a", "created_at": "2026-11-28T06:17:21.283504+00:00", "elapsed_ms": 35.021, "execution_result": {"artifact_id": "ws_20261128_c44855cfb6f1", "columns": ["game_id", "white_player", "black_player", "result", "white_elo", "black_elo", "game_timestamp", "eval_cp_white", "san_by_ply"], "is_complete": true, "is_truncated": false, "result_type": "artifact", "row_count": 0, "rows": [], "scalar_value": null}, "format": "parquet", "is_complete": true, "is_truncated": false, "kind": "tabl …[truncated for display]
  artifacts: {"artifact_metadata": {"artifact_complete": true, "columns": [{"name": "game_id", "type": "null"}, {"name": "white_player", "type": "null"}, {"name": "black_player", "type": "null"}, {"name": "result", "type": "null"}, {"name": "white_elo", "type": "null"}, {"name": "black_elo", "type": "null"}, {"name": "game_timestamp", "type": "null"}, {"name": "eval_cp_white", "type": "null"}, {"name": "san_by_ply", "type": "null"}], "complete": true, "row_count": 0}, "artifact_references": {"artifact_id": "ws_20261128_c44855cfb6f1", "parent_artifact_ids": []}}
Iteration 3 (? ms): V2 deterministic tool selection
Tool step_3: orient_game_rows
  arguments: {"player_color": "${inputs.player_color}", "player_identifier": "${steps.step_1.scalar_value}", "source_artifact": "${steps.step_2.artifact_id}", "summary": "Games oriented to the selected player's perspective"}
  resolved_arguments: {"player_color": "black", "player_identifier": "my_name", "source_artifact": "ws_20261128_c44855cfb6f1", "summary": "Games oriented to the selected player's perspective"}
  result: {"artifact_complete": true, "artifact_id": "ws_20261128_19c5db25dc3c", "columns": [{"name": "game_id", "type": "null"}, {"name": "white_player", "type": "null"}, {"name": "black_player", "type": "null"}, {"name": "result", "type": "null"}, {"name": "white_elo", "type": "null"}, {"name": "black_elo", "type": "null"}, {"name": "game_timestamp", "type": "null"}, {"name": "eval_cp_white", "type": "null"}, {"name": "san_by_ply", "type": "null"}, {"name": "player_color", "type": "string"}, {"name": "player_name", "type": "string"}, {"name": "opponent_name", "type": "string"}, {"name": "player_score", "type": "double"}, {"name": "player_result", "type": "string"}, {"name": "player_rating", "type": "null"}, {"name": "opponent_rating", "type": "null"}, {"name": "player_eval_cp", "type": "double"}], "columns_preserved": ["game_id", "white_player", "black_player", "result", "white_elo", "black_elo", "game_timestamp", "eval_cp_white", "san_by_ply"], "excluded_color_rows": 0, "excluded_nonparticipant_rows": 0, "normalized_columns": ["player_color", "player_name", "opponent_name", "player_score", "player_result", "player_rating", "opponent_rating", "player_eval_cp"], "oriented_rows": 0, "query_t …[truncated for display]
  artifacts: {"artifact_metadata": {"artifact_complete": true, "columns": [{"name": "game_id", "type": "null"}, {"name": "white_player", "type": "null"}, {"name": "black_player", "type": "null"}, {"name": "result", "type": "null"}, {"name": "white_elo", "type": "null"}, {"name": "black_elo", "type": "null"}, {"name": "game_timestamp", "type": "null"}, {"name": "eval_cp_white", "type": "null"}, {"name": "san_by_ply", "type": "null"}, {"name": "player_color", "type": "string"}, {"name": "player_name", "type": "string"}, {"name": "opponent_name", "type": "string"}, {"name": "player_score", "type": "double"}, {"name": "player_result", "type": "string"}, {"name": "player_rating", "type": "null"}, {"name": "opponent_rating", "type": "null"}, {"name": "player_eval_cp", "type": "double"}], "row_count": 0}, "artifact_references": {"artifact_id": "ws_20261128_19c5db25dc3c", "source_artifact_id": "ws_20261128_c44855cfb6f1"}}
Iteration 4 (? ms): V2 deterministic tool selection
Tool step_4: query_workspace_readonly
  arguments: {"max_inline_rows": 1, "parameters": {"ply_count": "${inputs.ply_count}"}, "result_mode": "inline", "sources": {"oriented": "${steps.step_3.artifact_id}"}, "sql": "WITH RECURSIVE projection(ply, san_columns) AS (SELECT 0, '' UNION ALL SELECT ply + 1, san_columns || printf(', json_extract(san_by_ply, ''$.\"%d\"'') AS Ply_%d', ply + 1, ply + 1) FROM projection WHERE ply < :ply_count) SELECT 'WITH chronological AS (SELECT *, row_number() OVER (ORDER BY game_timestamp, game_id) AS game_sequence FROM oriented), first_seen AS (SELECT *, min(game_sequence) OVER (PARTITION BY opponent_name) AS first_opponent_sequence FROM chronological), numbered AS (SELECT *, dense_rank() OVER (ORDER BY CASE WHEN opponent_name IS NULL THEN 1 ELSE 0 END, first_opponent_sequence) AS opponent_number FROM first_seen) SELECT player_score AS Outcome, game_timestamp AS Time' || san_columns || ', player_eval_cp AS Evaluation_cp, CASE WHEN opponent_name IS NOT NULL THEN printf(''opponent_%03d'', opponent_number) END AS Opponent, opponent_rating AS Opponent_rating FROM numbered ORDER BY game_sequence' AS dataset_query FROM projection WHERE ply = :ply_count"}
  resolved_arguments: {"max_inline_rows": 1, "parameters": {"ply_count": 10}, "result_mode": "inline", "sources": {"oriented": "ws_20261128_19c5db25dc3c"}, "sql": "WITH RECURSIVE projection(ply, san_columns) AS (SELECT 0, '' UNION ALL SELECT ply + 1, san_columns || printf(', json_extract(san_by_ply, ''$.\"%d\"'') AS Ply_%d', ply + 1, ply + 1) FROM projection WHERE ply < :ply_count) SELECT 'WITH chronological AS (SELECT *, row_number() OVER (ORDER BY game_timestamp, game_id) AS game_sequence FROM oriented), first_seen AS (SELECT *, min(game_sequence) OVER (PARTITION BY opponent_name) AS first_opponent_sequence FROM chronological), numbered AS (SELECT *, dense_rank() OVER (ORDER BY CASE WHEN opponent_name IS NULL THEN 1 ELSE 0 END, first_opponent_sequence) AS opponent_number FROM first_seen) SELECT player_score AS Outcome, game_timestamp AS Time' || san_columns || ', player_eval_cp AS Evaluation_cp, CASE WHEN opponent_name IS NOT NULL THEN printf(''opponent_%03d'', opponent_number) END AS Opponent, opponent_rating AS Opponent_rating FROM numbered ORDER BY game_sequence' AS dataset_query FROM projection WHERE ply = :ply_count"}
  result: {"artifact_complete": false, "complete": false, "elapsed_ms": 5.087, "error": {"category": "unknown_relation", "message": "Workspace SQL may reference only supplied source aliases; unknown relation: projection"}, "query_truncated": false, "status": "error"}
  artifacts: {"artifact_metadata": {"artifact_complete": false, "complete": false}, "artifact_references": {}}
Iteration 5 (? ms): V2 deterministic tool selection
Tool step_5: query_workspace_readonly
  arguments: {"max_inline_rows": 1, "parameters": {"ply_count": "${inputs.ply_count}"}, "result_mode": "inline", "sources": {"oriented": "${steps.step_3.artifact_id}"}, "sql": "WITH digits(n) AS (VALUES (0),(1),(2),(3),(4),(5),(6),(7),(8),(9)), plies AS (SELECT 1 + a.n + 10*b.n + 100*c.n + 1000*d.n AS ply FROM digits a CROSS JOIN digits b CROSS JOIN digits c CROSS JOIN digits d), pieces AS (SELECT printf(', json_extract(san_by_ply, ''$.\"%d\"'') AS Ply_%d', ply, ply) AS fragment FROM plies WHERE ply <= :ply_count ORDER BY ply) SELECT 'WITH chronological AS (SELECT *, row_number() OVER (ORDER BY game_timestamp, game_id) AS game_sequence FROM oriented), first_seen AS (SELECT *, min(game_sequence) OVER (PARTITION BY opponent_name) AS first_opponent_sequence FROM chronological), numbered AS (SELECT *, dense_rank() OVER (ORDER BY CASE WHEN opponent_name IS NULL THEN 1 ELSE 0 END, first_opponent_sequence) AS opponent_number FROM first_seen) SELECT player_score AS Outcome, game_timestamp AS Time' || coalesce(group_concat(fragment, ''), '') || ', player_eval_cp AS Evaluation_cp, CASE WHEN opponent_name IS NOT NULL THEN printf(''opponent_%03d'', opponent_number) END AS Opponent, opponent_rating AS Opponent_rating FROM numbered ORDER BY game_sequence' AS dataset_query FROM pieces"}
  resolved_arguments: {"max_inline_rows": 1, "parameters": {"ply_count": 10}, "result_mode": "inline", "sources": {"oriented": "ws_20261128_19c5db25dc3c"}, "sql": "WITH digits(n) AS (VALUES (0),(1),(2),(3),(4),(5),(6),(7),(8),(9)), plies AS (SELECT 1 + a.n + 10*b.n + 100*c.n + 1000*d.n AS ply FROM digits a CROSS JOIN digits b CROSS JOIN digits c CROSS JOIN digits d), pieces AS (SELECT printf(', json_extract(san_by_ply, ''$.\"%d\"'') AS Ply_%d', ply, ply) AS fragment FROM plies WHERE ply <= :ply_count ORDER BY ply) SELECT 'WITH chronological AS (SELECT *, row_number() OVER (ORDER BY game_timestamp, game_id) AS game_sequence FROM oriented), first_seen AS (SELECT *, min(game_sequence) OVER (PARTITION BY opponent_name) AS first_opponent_sequence FROM chronological), numbered AS (SELECT *, dense_rank() OVER (ORDER BY CASE WHEN opponent_name IS NULL THEN 1 ELSE 0 END, first_opponent_sequence) AS opponent_number FROM first_seen) SELECT player_score AS Outcome, game_timestamp AS Time' || coalesce(group_concat(fragment, ''), '') || ', player_eval_cp AS Evaluation_cp, CASE WHEN opponent_name IS NOT NULL THEN printf(''opponent_%03d'', opponent_number) END AS Opponent, opponent_rating AS Opponent_rating FROM numbered ORDER BY game_sequence' AS dataset_query FROM pieces"}
  result: {"artifact_complete": false, "complete": false, "elapsed_ms": 3.165, "error": {"category": "unknown_relation", "message": "Workspace SQL may reference only supplied source aliases; unknown relation: digits"}, "query_truncated": false, "status": "error"}
  artifacts: {"artifact_metadata": {"artifact_complete": false, "complete": false}, "artifact_references": {}}
Iteration 6 (? ms): V2 deterministic tool selection
Tool step_6: query_database_readonly
  arguments: {"max_rows": 1, "parameters": {"ply_count": "${inputs.ply_count}"}, "sql": "WITH RECURSIVE projection(ply, san_columns) AS (SELECT 0, '' UNION ALL SELECT ply + 1, san_columns || printf(', json_extract(san_by_ply, ''$.\"%d\"'') AS Ply_%d', ply + 1, ply + 1) FROM projection WHERE ply < :ply_count) SELECT 'WITH chronological AS (SELECT *, row_number() OVER (ORDER BY game_timestamp, game_id) AS game_sequence FROM oriented), first_seen AS (SELECT *, min(game_sequence) OVER (PARTITION BY opponent_name) AS first_opponent_sequence FROM chronological), numbered AS (SELECT *, dense_rank() OVER (ORDER BY CASE WHEN opponent_name IS NULL THEN 1 ELSE 0 END, first_opponent_sequence) AS opponent_number FROM first_seen) SELECT player_score AS Outcome, game_timestamp AS Time' || san_columns || ', player_eval_cp AS Evaluation_cp, CASE WHEN opponent_name IS NOT NULL THEN printf(''opponent_%03d'', opponent_number) END AS Opponent, opponent_rating AS Opponent_rating FROM numbered ORDER BY game_sequence' AS dataset_query FROM projection WHERE ply = :ply_count"}
  resolved_arguments: {"max_rows": 1, "parameters": {"ply_count": 10}, "sql": "WITH RECURSIVE projection(ply, san_columns) AS (SELECT 0, '' UNION ALL SELECT ply + 1, san_columns || printf(', json_extract(san_by_ply, ''$.\"%d\"'') AS Ply_%d', ply + 1, ply + 1) FROM projection WHERE ply < :ply_count) SELECT 'WITH chronological AS (SELECT *, row_number() OVER (ORDER BY game_timestamp, game_id) AS game_sequence FROM oriented), first_seen AS (SELECT *, min(game_sequence) OVER (PARTITION BY opponent_name) AS first_opponent_sequence FROM chronological), numbered AS (SELECT *, dense_rank() OVER (ORDER BY CASE WHEN opponent_name IS NULL THEN 1 ELSE 0 END, first_opponent_sequence) AS opponent_number FROM first_seen) SELECT player_score AS Outcome, game_timestamp AS Time' || san_columns || ', player_eval_cp AS Evaluation_cp, CASE WHEN opponent_name IS NOT NULL THEN printf(''opponent_%03d'', opponent_number) END AS Opponent, opponent_rating AS Opponent_rating FROM numbered ORDER BY game_sequence' AS dataset_query FROM projection WHERE ply = :ply_count"}
  result: {"columns": ["dataset_query"], "complete": true, "effective_max_rows": 1, "elapsed_ms": 0.605, "execution_result": {"artifact_id": null, "columns": ["dataset_query"], "is_complete": true, "is_truncated": false, "result_type": "scalar", "row_count": 1, "rows": [{"dataset_query": "WITH chronological AS (SELECT *, row_number() OVER (ORDER BY game_timestamp, game_id) AS game_sequence FROM oriented), first_seen AS (SELECT *, min(game_sequence) OVER (PARTITION BY opponent_name) AS first_opponent_sequence FROM chronological), numbered AS (SELECT *, dense_rank() OVER (ORDER BY CASE WHEN opponent_name IS NULL THEN 1 ELSE 0 END, first_opponent_sequence) AS opponent_number FROM first_seen) SELECT player_score AS Outcome, game_timestamp AS Time, json_extract(san_by_ply, '$.\"1\"') AS Ply_1, json_extract(san_by_ply, '$.\"2\"') AS Ply_2, json_extract(san_by_ply, '$.\"3\"') AS Ply_3, json_extract(san_by_ply, '$.\"4\"') AS Ply_4, json_extract(san_by_ply, '$.\"5\"') AS Ply_5, json_extract(san_by_ply, '$.\"6\"') AS Ply_6, json_extract(san_by_ply, '$.\"7\"') AS Ply_7, json_extract(san_by_ply, '$.\"8\"') AS Ply_8, json_extract(san_by_ply, '$.\"9\"') AS Ply_9, json_extract(san_by_ply, '$.\"10\"') AS Pl …[truncated for display]
  artifacts: {"artifact_metadata": {"columns": ["dataset_query"], "complete": true, "row_count": 1}, "artifact_references": {}}
Iteration 7 (? ms): V2 deterministic tool selection
Tool step_7: query_workspace_readonly
  arguments: {"result_mode": "workspace", "sources": {"oriented": "${steps.step_3.artifact_id}"}, "sql": "${steps.step_6.scalar_value}"}
  resolved_arguments: {"result_mode": "workspace", "sources": {"oriented": "ws_20261128_19c5db25dc3c"}, "sql": "WITH chronological AS (SELECT *, row_number() OVER (ORDER BY game_timestamp, game_id) AS game_sequence FROM oriented), first_seen AS (SELECT *, min(game_sequence) OVER (PARTITION BY opponent_name) AS first_opponent_sequence FROM chronological), numbered AS (SELECT *, dense_rank() OVER (ORDER BY CASE WHEN opponent_name IS NULL THEN 1 ELSE 0 END, first_opponent_sequence) AS opponent_number FROM first_seen) SELECT player_score AS Outcome, game_timestamp AS Time, json_extract(san_by_ply, '$.\"1\"') AS Ply_1, json_extract(san_by_ply, '$.\"2\"') AS Ply_2, json_extract(san_by_ply, '$.\"3\"') AS Ply_3, json_extract(san_by_ply, '$.\"4\"') AS Ply_4, json_extract(san_by_ply, '$.\"5\"') AS Ply_5, json_extract(san_by_ply, '$.\"6\"') AS Ply_6, json_extract(san_by_ply, '$.\"7\"') AS Ply_7, json_extract(san_by_ply, '$.\"8\"') AS Ply_8, json_extract(san_by_ply, '$.\"9\"') AS Ply_9, json_extract(san_by_ply, '$.\"10\"') AS Ply_10, player_eval_cp AS Evaluation_cp, CASE WHEN opponent_name IS NOT NULL THEN printf('opponent_%03d', opponent_number) END AS Opponent, opponent_rating AS Opponent_rating FROM numbered ORDER BY game_sequence"}
  result: {"artifact_complete": true, "artifact_id": "ws_20261128_5b3fa5ab3d80", "column_names": ["Outcome", "Time", "Ply_1", "Ply_2", "Ply_3", "Ply_4", "Ply_5", "Ply_6", "Ply_7", "Ply_8", "Ply_9", "Ply_10", "Evaluation_cp", "Opponent", "Opponent_rating"], "columns": [{"name": "Outcome", "type": "double"}, {"name": "Time", "type": "int32"}, {"name": "Ply_1", "type": "string"}, {"name": "Ply_2", "type": "string"}, {"name": "Ply_3", "type": "string"}, {"name": "Ply_4", "type": "string"}, {"name": "Ply_5", "type": "string"}, {"name": "Ply_6", "type": "string"}, {"name": "Ply_7", "type": "string"}, {"name": "Ply_8", "type": "string"}, {"name": "Ply_9", "type": "string"}, {"name": "Ply_10", "type": "string"}, {"name": "Evaluation_cp", "type": "double"}, {"name": "Opponent", "type": "string"}, {"name": "Opponent_rating", "type": "int32"}], "complete": true, "elapsed_ms": 44.871, "execution_result": {"artifact_id": "ws_20261128_5b3fa5ab3d80", "columns": ["Outcome", "Time", "Ply_1", "Ply_2", "Ply_3", "Ply_4", "Ply_5", "Ply_6", "Ply_7", "Ply_8", "Ply_9", "Ply_10", "Evaluation_cp", "Opponent", "Opponent_rating"], "is_complete": true, "is_truncated": false, "result_type": "artifact", "row_count": 0, "r …[truncated for display]
  artifacts: {"artifact_metadata": {"artifact_complete": true, "columns": [{"name": "Outcome", "type": "double"}, {"name": "Time", "type": "int32"}, {"name": "Ply_1", "type": "string"}, {"name": "Ply_2", "type": "string"}, {"name": "Ply_3", "type": "string"}, {"name": "Ply_4", "type": "string"}, {"name": "Ply_5", "type": "string"}, {"name": "Ply_6", "type": "string"}, {"name": "Ply_7", "type": "string"}, {"name": "Ply_8", "type": "string"}, {"name": "Ply_9", "type": "string"}, {"name": "Ply_10", "type": "string"}, {"name": "Evaluation_cp", "type": "double"}, {"name": "Opponent", "type": "string"}, {"name": "Opponent_rating", "type": "int32"}], "complete": true, "row_count": 0}, "artifact_references": {"artifact_id": "ws_20261128_5b3fa5ab3d80"}}
Iteration 8 (? ms): V2 deterministic tool selection
Tool step_8: export_workspace_dataset
  arguments: {"output_filename": "${inputs.output_filename}", "source_artifact": "${steps.step_7.artifact_id}"}
  resolved_arguments: {"output_filename": "research_table.csv", "source_artifact": "ws_20261128_5b3fa5ab3d80"}
  result: {"artifact_complete": true, "artifact_id": "ws_20261128_4a8d206af317", "content_hash": "647ab3edc511184b05826f783e1b7c0d7abd5b4c0b167da08c0c668251fbe4a9", "format": "csv", "output_filename": "research_table.csv", "row_count": 0, "source_artifact_id": "ws_20261128_5b3fa5ab3d80", "staged_effect_ids": ["analysis_08ca93ef4cfc46e384e3c916f2ad0cac:effect_1"], "staged_output_path": "research_table.csv", "status": "ok"}
  artifacts: {"artifact_metadata": {"artifact_complete": true, "row_count": 0}, "artifact_references": {"artifact_id": "ws_20261128_4a8d206af317", "source_artifact_id": "ws_20261128_5b3fa5ab3d80"}}
Iteration 9 (? ms): V2 deterministic tool selection
Tool step_9: present_scalar
  arguments: {"value": "${steps.step_7.row_count}"}
  resolved_arguments: {"value": 0}
  result: {"rendered_output": "0", "result_type": "scalar", "scalar_value": 0, "staged_effect_ids": ["analysis_08ca93ef4cfc46e384e3c916f2ad0cac:effect_2"], "status": "ok"}
  artifacts: {"artifact_metadata": {}, "artifact_references": {}}
Iteration 10 (? ms): V2 deterministic tool selection
Tool step_10: query_workspace_readonly
  arguments: {"parameters": {"preview_count": "${inputs.preview_row_count}"}, "result_mode": "workspace", "sources": {"dataset": "${steps.step_7.artifact_id}"}, "sql": "SELECT * FROM dataset ORDER BY rowid LIMIT :preview_count"}
  resolved_arguments: {"parameters": {"preview_count": 5}, "result_mode": "workspace", "sources": {"dataset": "ws_20261128_5b3fa5ab3d80"}, "sql": "SELECT * FROM dataset ORDER BY rowid LIMIT :preview_count"}
  result: {"artifact_complete": false, "complete": false, "elapsed_ms": 19.064, "error": {"category": "invalid_workspace_column", "message": "Invalid workspace query: column \"rowid\" does not exist in the supplied source artifact schema. Available columns by source:\n  dataset (ws_20261128_5b3fa5ab3d80): Outcome, Time, Ply_1, Ply_2, Ply_3, Ply_4, Ply_5, Ply_6, Ply_7, Ply_8, Ply_9, Ply_10, Evaluation_cp, Opponent, Opponent_rating"}, "query_truncated": false, "status": "error"}
  artifacts: {"artifact_metadata": {"artifact_complete": false, "complete": false}, "artifact_references": {}}
Iteration 11 (? ms): V2 deterministic tool selection
Tool step_11: query_database_readonly
  arguments: {"max_rows": 1, "parameters": {"dataset_query": "${steps.step_6.scalar_value}", "preview_count": "${inputs.preview_row_count}"}, "sql": "SELECT :dataset_query || ' LIMIT ' || printf('%d', :preview_count) AS preview_query"}
  resolved_arguments: {"max_rows": 1, "parameters": {"dataset_query": "WITH chronological AS (SELECT *, row_number() OVER (ORDER BY game_timestamp, game_id) AS game_sequence FROM oriented), first_seen AS (SELECT *, min(game_sequence) OVER (PARTITION BY opponent_name) AS first_opponent_sequence FROM chronological), numbered AS (SELECT *, dense_rank() OVER (ORDER BY CASE WHEN opponent_name IS NULL THEN 1 ELSE 0 END, first_opponent_sequence) AS opponent_number FROM first_seen) SELECT player_score AS Outcome, game_timestamp AS Time, json_extract(san_by_ply, '$.\"1\"') AS Ply_1, json_extract(san_by_ply, '$.\"2\"') AS Ply_2, json_extract(san_by_ply, '$.\"3\"') AS Ply_3, json_extract(san_by_ply, '$.\"4\"') AS Ply_4, json_extract(san_by_ply, '$.\"5\"') AS Ply_5, json_extract(san_by_ply, '$.\"6\"') AS Ply_6, json_extract(san_by_ply, '$.\"7\"') AS Ply_7, json_extract(san_by_ply, '$.\"8\"') AS Ply_8, json_extract(san_by_ply, '$.\"9\"') AS Ply_9, json_extract(san_by_ply, '$.\"10\"') AS Ply_10, player_eval_cp AS Evaluation_cp, CASE WHEN opponent_name IS NOT NULL THEN printf('opponent_%03d', opponent_number) END AS Opponent, opponent_rating AS Opponent_rating FROM numbered ORDER BY game_sequence", "preview_count": 5}, "sql": "SELECT :dataset_query || ' LIMIT ' || printf('%d', :preview_count) AS preview_query"}
  result: {"columns": ["preview_query"], "complete": true, "effective_max_rows": 1, "elapsed_ms": 0.517, "execution_result": {"artifact_id": null, "columns": ["preview_query"], "is_complete": true, "is_truncated": false, "result_type": "scalar", "row_count": 1, "rows": [{"preview_query": "WITH chronological AS (SELECT *, row_number() OVER (ORDER BY game_timestamp, game_id) AS game_sequence FROM oriented), first_seen AS (SELECT *, min(game_sequence) OVER (PARTITION BY opponent_name) AS first_opponent_sequence FROM chronological), numbered AS (SELECT *, dense_rank() OVER (ORDER BY CASE WHEN opponent_name IS NULL THEN 1 ELSE 0 END, first_opponent_sequence) AS opponent_number FROM first_seen) SELECT player_score AS Outcome, game_timestamp AS Time, json_extract(san_by_ply, '$.\"1\"') AS Ply_1, json_extract(san_by_ply, '$.\"2\"') AS Ply_2, json_extract(san_by_ply, '$.\"3\"') AS Ply_3, json_extract(san_by_ply, '$.\"4\"') AS Ply_4, json_extract(san_by_ply, '$.\"5\"') AS Ply_5, json_extract(san_by_ply, '$.\"6\"') AS Ply_6, json_extract(san_by_ply, '$.\"7\"') AS Ply_7, json_extract(san_by_ply, '$.\"8\"') AS Ply_8, json_extract(san_by_ply, '$.\"9\"') AS Ply_9, json_extract(san_by_ply, '$.\"10\"') AS Pl …[truncated for display]
  artifacts: {"artifact_metadata": {"columns": ["preview_query"], "complete": true, "row_count": 1}, "artifact_references": {}}
Iteration 12 (? ms): V2 deterministic tool selection
Tool step_12: query_database_readonly
  arguments: {"max_rows": 1, "parameters": {"dataset_query": "${steps.step_6.scalar_value}"}, "sql": "SELECT :dataset_query || ' LIMIT :preview_count' AS preview_query"}
  resolved_arguments: {"max_rows": 1, "parameters": {"dataset_query": "WITH chronological AS (SELECT *, row_number() OVER (ORDER BY game_timestamp, game_id) AS game_sequence FROM oriented), first_seen AS (SELECT *, min(game_sequence) OVER (PARTITION BY opponent_name) AS first_opponent_sequence FROM chronological), numbered AS (SELECT *, dense_rank() OVER (ORDER BY CASE WHEN opponent_name IS NULL THEN 1 ELSE 0 END, first_opponent_sequence) AS opponent_number FROM first_seen) SELECT player_score AS Outcome, game_timestamp AS Time, json_extract(san_by_ply, '$.\"1\"') AS Ply_1, json_extract(san_by_ply, '$.\"2\"') AS Ply_2, json_extract(san_by_ply, '$.\"3\"') AS Ply_3, json_extract(san_by_ply, '$.\"4\"') AS Ply_4, json_extract(san_by_ply, '$.\"5\"') AS Ply_5, json_extract(san_by_ply, '$.\"6\"') AS Ply_6, json_extract(san_by_ply, '$.\"7\"') AS Ply_7, json_extract(san_by_ply, '$.\"8\"') AS Ply_8, json_extract(san_by_ply, '$.\"9\"') AS Ply_9, json_extract(san_by_ply, '$.\"10\"') AS Ply_10, player_eval_cp AS Evaluation_cp, CASE WHEN opponent_name IS NOT NULL THEN printf('opponent_%03d', opponent_number) END AS Opponent, opponent_rating AS Opponent_rating FROM numbered ORDER BY game_sequence"}, "sql": "SELECT :dataset_query || ' LIMIT :preview_count' AS preview_query"}
  result: {"columns": ["preview_query"], "complete": true, "effective_max_rows": 1, "elapsed_ms": 0.376, "execution_result": {"artifact_id": null, "columns": ["preview_query"], "is_complete": true, "is_truncated": false, "result_type": "scalar", "row_count": 1, "rows": [{"preview_query": "WITH chronological AS (SELECT *, row_number() OVER (ORDER BY game_timestamp, game_id) AS game_sequence FROM oriented), first_seen AS (SELECT *, min(game_sequence) OVER (PARTITION BY opponent_name) AS first_opponent_sequence FROM chronological), numbered AS (SELECT *, dense_rank() OVER (ORDER BY CASE WHEN opponent_name IS NULL THEN 1 ELSE 0 END, first_opponent_sequence) AS opponent_number FROM first_seen) SELECT player_score AS Outcome, game_timestamp AS Time, json_extract(san_by_ply, '$.\"1\"') AS Ply_1, json_extract(san_by_ply, '$.\"2\"') AS Ply_2, json_extract(san_by_ply, '$.\"3\"') AS Ply_3, json_extract(san_by_ply, '$.\"4\"') AS Ply_4, json_extract(san_by_ply, '$.\"5\"') AS Ply_5, json_extract(san_by_ply, '$.\"6\"') AS Ply_6, json_extract(san_by_ply, '$.\"7\"') AS Ply_7, json_extract(san_by_ply, '$.\"8\"') AS Ply_8, json_extract(san_by_ply, '$.\"9\"') AS Ply_9, json_extract(san_by_ply, '$.\"10\"') AS Pl …[truncated for display]
  artifacts: {"artifact_metadata": {"columns": ["preview_query"], "complete": true, "row_count": 1}, "artifact_references": {}}
Iteration 13 (? ms): V2 deterministic tool selection
Tool step_13: query_workspace_readonly
  arguments: {"parameters": {"preview_count": "${inputs.preview_row_count}"}, "result_mode": "workspace", "sources": {"oriented": "${steps.step_3.artifact_id}"}, "sql": "${steps.step_12.scalar_value}"}
  resolved_arguments: {"parameters": {"preview_count": 5}, "result_mode": "workspace", "sources": {"oriented": "ws_20261128_19c5db25dc3c"}, "sql": "WITH chronological AS (SELECT *, row_number() OVER (ORDER BY game_timestamp, game_id) AS game_sequence FROM oriented), first_seen AS (SELECT *, min(game_sequence) OVER (PARTITION BY opponent_name) AS first_opponent_sequence FROM chronological), numbered AS (SELECT *, dense_rank() OVER (ORDER BY CASE WHEN opponent_name IS NULL THEN 1 ELSE 0 END, first_opponent_sequence) AS opponent_number FROM first_seen) SELECT player_score AS Outcome, game_timestamp AS Time, json_extract(san_by_ply, '$.\"1\"') AS Ply_1, json_extract(san_by_ply, '$.\"2\"') AS Ply_2, json_extract(san_by_ply, '$.\"3\"') AS Ply_3, json_extract(san_by_ply, '$.\"4\"') AS Ply_4, json_extract(san_by_ply, '$.\"5\"') AS Ply_5, json_extract(san_by_ply, '$.\"6\"') AS Ply_6, json_extract(san_by_ply, '$.\"7\"') AS Ply_7, json_extract(san_by_ply, '$.\"8\"') AS Ply_8, json_extract(san_by_ply, '$.\"9\"') AS Ply_9, json_extract(san_by_ply, '$.\"10\"') AS Ply_10, player_eval_cp AS Evaluation_cp, CASE WHEN opponent_name IS NOT NULL THEN printf('opponent_%03d', opponent_number) END AS Opponent, opponent_rating AS Opponent_rating FROM numbered ORDER BY game_sequence LIMIT :preview_count"}
  result: {"artifact_complete": true, "artifact_id": "ws_20261128_dea1b8f070dd", "column_names": ["Outcome", "Time", "Ply_1", "Ply_2", "Ply_3", "Ply_4", "Ply_5", "Ply_6", "Ply_7", "Ply_8", "Ply_9", "Ply_10", "Evaluation_cp", "Opponent", "Opponent_rating"], "columns": [{"name": "Outcome", "type": "double"}, {"name": "Time", "type": "int32"}, {"name": "Ply_1", "type": "string"}, {"name": "Ply_2", "type": "string"}, {"name": "Ply_3", "type": "string"}, {"name": "Ply_4", "type": "string"}, {"name": "Ply_5", "type": "string"}, {"name": "Ply_6", "type": "string"}, {"name": "Ply_7", "type": "string"}, {"name": "Ply_8", "type": "string"}, {"name": "Ply_9", "type": "string"}, {"name": "Ply_10", "type": "string"}, {"name": "Evaluation_cp", "type": "double"}, {"name": "Opponent", "type": "string"}, {"name": "Opponent_rating", "type": "int32"}], "complete": true, "elapsed_ms": 24.643, "execution_result": {"artifact_id": "ws_20261128_dea1b8f070dd", "columns": ["Outcome", "Time", "Ply_1", "Ply_2", "Ply_3", "Ply_4", "Ply_5", "Ply_6", "Ply_7", "Ply_8", "Ply_9", "Ply_10", "Evaluation_cp", "Opponent", "Opponent_rating"], "is_complete": true, "is_truncated": false, "result_type": "artifact", "row_count": 0, "r …[truncated for display]
  artifacts: {"artifact_metadata": {"artifact_complete": true, "columns": [{"name": "Outcome", "type": "double"}, {"name": "Time", "type": "int32"}, {"name": "Ply_1", "type": "string"}, {"name": "Ply_2", "type": "string"}, {"name": "Ply_3", "type": "string"}, {"name": "Ply_4", "type": "string"}, {"name": "Ply_5", "type": "string"}, {"name": "Ply_6", "type": "string"}, {"name": "Ply_7", "type": "string"}, {"name": "Ply_8", "type": "string"}, {"name": "Ply_9", "type": "string"}, {"name": "Ply_10", "type": "string"}, {"name": "Evaluation_cp", "type": "double"}, {"name": "Opponent", "type": "string"}, {"name": "Opponent_rating", "type": "int32"}], "complete": true, "row_count": 0}, "artifact_references": {"artifact_id": "ws_20261128_dea1b8f070dd"}}
Iteration 14 (? ms): V2 deterministic tool selection
Tool step_14: present_table
  arguments: {"source_artifact": "${steps.step_13.artifact_id}"}
  resolved_arguments: {"source_artifact": "ws_20261128_dea1b8f070dd"}
  result: {"columns": ["Outcome", "Time", "Ply_1", "Ply_2", "Ply_3", "Ply_4", "Ply_5", "Ply_6", "Ply_7", "Ply_8", "Ply_9", "Ply_10", "Evaluation_cp", "Opponent", "Opponent_rating"], "complete": true, "rendered_output": "Outcome | Time | Ply_1 | Ply_2 | Ply_3 | Ply_4 | Ply_5 | Ply_6 | Ply_7 | Ply_8 | Ply_9 | Ply_10 | Evaluation_cp | Opponent | Opponent_rating\n--- | --- | --- | --- | --- | --- | --- | --- | --- | --- | --- | --- | --- | --- | ---", "result_type": "table", "row_count": 0, "source_artifact_id": "ws_20261128_dea1b8f070dd", "staged_effect_ids": ["analysis_08ca93ef4cfc46e384e3c916f2ad0cac:effect_3"], "status": "ok"}
  artifacts: {"artifact_metadata": {"columns": ["Outcome", "Time", "Ply_1", "Ply_2", "Ply_3", "Ply_4", "Ply_5", "Ply_6", "Ply_7", "Ply_8", "Ply_9", "Ply_10", "Evaluation_cp", "Opponent", "Opponent_rating"], "complete": true, "row_count": 0}, "artifact_references": {"source_artifact_id": "ws_20261128_dea1b8f070dd"}}

\end{Verbatim}

\subsection{Final pipeline after the LLM3 revision}

The following listing reproduces the final pipeline after the LLM3 revision.
In final step~2, the generated SAN expressions
wrap \texttt{json\_extract} in \texttt{trim} to remove surrounding double quotes.
The canonical task includes the corresponding instruction. After input-reference
conversion and step renumbering, the other retained tool arguments match the
discovery record. The output declarations select final steps~8, 9, and~10.

The stored defaults remain $N=10$ and the period from 1 January 2024 through
31 December 2025. The later run with $N=4$ and a start date of 1 January 2025
used parameter overrides, which are not stored as new defaults in this artifact.

\begin{Verbatim}
name: t
canonical_task: 'Use the chess research database to select blitz games in which ${inputs.player_identifier}
  played as ${inputs.player_color}. Include games played from ${inputs.start_date}
  through ${inputs.end_date}, inclusive, with at least ${inputs.ply_count} plies.
  Produce one row per qualifying game, sorted by game timestamp from oldest to newest,
  with a deterministic tie-breaker for identical timestamps.


  Use these columns in order: (1) outcome from the selected player''s perspective:
  win = 1, draw = 0.5, loss = 0, and missing or unrecognized results left blank; (2)
  Time, the recorded game timestamp; (3) one column per ply for the first ${inputs.ply_count}
  plies, in play order and in SAN without surrounding double quotes; (4) the stored
  evaluation of the position immediately after ply ${inputs.ply_count}, expressed
  from the selected player''s perspective; (5) anonymized opponent name; (6) the opponent''s
  recorded rating for that game. For the evaluation, reverse the sign of a documented
  White-relative value only when the selected player is Black, and do not negate a
  value already expressed from the selected player''s perspective. Use the stored
  evaluation as recorded without adding engine-run or depth restrictions. Leave missing
  evaluations and ratings blank rather than imputing them.


  Assign opponent aliases sequentially in order of first appearance in the sorted
  dataset: opponent_001, opponent_002, and so on. Reuse the same alias for the same
  opponent across games. Do not expose original opponent names or the alias mapping
  in either output.


  Write the complete sorted dataset to ${inputs.output_filename}. Separately, on screen,
  report the total number of game rows, excluding the header, and display only the
  first ${inputs.preview_row_count} rows of the same sorted dataset, or all rows if
  fewer qualify. Do not add other user-facing outputs.'
canonical_inputs:
  player_identifier:
    type: string
    value: my_name
  player_color:
    type: string
    value: black
  time_control:
    type: string
    value: blitz
  start_date:
    type: string
    value: 1. jan 2024
  end_date:
    type: string
    value: 31. dec 2025
  ply_count:
    type: integer
    value: 10
  output_filename:
    type: string
    value: research_table.csv
  preview_row_count:
    type: integer
    value: 5
parameters:
  player_identifier:
    type: string
    default: my_name
  ply_count:
    type: integer
    default: 10
  start_date:
    type: string
    default: 1. jan 2024
  end_date:
    type: string
    default: 31. dec 2025
  time_control:
    type: string
    default: blitz
  player_color:
    type: string
    default: black
  preview_row_count:
    type: integer
    default: 5
  output_filename:
    type: string
    default: research_table.csv
steps:
- id: step_1
  tool: resolve_player_identifier
  arguments:
    player_identifier: ${parameters.player_identifier}
- id: step_2
  tool: query_database_readonly
  arguments:
    sql: WITH RECURSIVE projection(ply, san_columns) AS (SELECT 0, '' UNION ALL SELECT
      ply + 1, san_columns || printf(', trim(json_extract(san_by_ply, ''$."%d"''),
      ''"'') AS Ply_%d', ply + 1, ply + 1) FROM projection WHERE ply < :ply_count)
      SELECT 'WITH chronological AS (SELECT *, row_number() OVER (ORDER BY game_timestamp,
      game_id) AS game_sequence FROM oriented), first_seen AS (SELECT *, min(game_sequence)
      OVER (PARTITION BY opponent_name) AS first_opponent_sequence FROM chronological),
      numbered AS (SELECT *, dense_rank() OVER (ORDER BY CASE WHEN opponent_name IS
      NULL THEN 1 ELSE 0 END, first_opponent_sequence) AS opponent_number FROM first_seen)
      SELECT player_score AS Outcome, game_timestamp AS Time' || san_columns || ',
      player_eval_cp AS Evaluation_cp, CASE WHEN opponent_name IS NOT NULL THEN printf(''opponent_%03d'',
      opponent_number) END AS Opponent, opponent_rating AS Opponent_rating FROM numbered
      ORDER BY game_sequence' AS dataset_query FROM projection WHERE ply = :ply_count
    parameters:
      ply_count: ${parameters.ply_count}
    max_rows: 1
- id: step_3
  tool: query_database_to_workspace
  arguments:
    sql: WITH date_inputs(kind, raw_date) AS (VALUES ('start', :start_date), ('end',
      :end_date)), date_parts AS (SELECT kind, trim(raw_date) AS raw_date, instr(trim(raw_date),
      '.') AS dot_pos FROM date_inputs), date_limits AS (SELECT kind, coalesce(date(replace(raw_date,
      '.', '-')), date(printf('%04d-%02d-%02d', cast(substr(raw_date, -4) AS integer),
      CASE lower(substr(trim(substr(raw_date, dot_pos + 1)), 1, 3)) WHEN 'jan' THEN
      1 WHEN 'feb' THEN 2 WHEN 'mar' THEN 3 WHEN 'apr' THEN 4 WHEN 'may' THEN 5 WHEN
      'jun' THEN 6 WHEN 'jul' THEN 7 WHEN 'aug' THEN 8 WHEN 'sep' THEN 9 WHEN 'oct'
      THEN 10 WHEN 'nov' THEN 11 WHEN 'dec' THEN 12 END, cast(substr(raw_date, 1,
      dot_pos - 1) AS integer)))) AS iso_day FROM date_parts), candidate_games AS
      (SELECT g.game_id, g.white_player, g.black_player, g.result, g.white_elo, g.black_elo,
      coalesce(date(replace(g.utc_date, '.', '-')), date(replace(g.date, '.', '-')))
      AS game_day, CASE WHEN time(g.utc_time) IS NOT NULL THEN g.utc_time END AS game_time
      FROM games g WHERE g.time_control_category = :speed AND ((:side = 'white' AND
      g.white_player = :player) OR (:side = 'black' AND g.black_player = :player)))
      SELECT g.game_id, g.white_player, g.black_player, g.result, g.white_elo, g.black_elo,
      CASE WHEN g.game_time IS NULL THEN g.game_day ELSE g.game_day || ' ' || g.game_time
      END AS game_timestamp, pos.eval_cp_white, (SELECT json_group_object(m.ply, m.san_move_played)
      FROM positions m WHERE m.game_id = g.game_id AND m.ply BETWEEN 1 AND :ply_count)
      AS san_by_ply FROM candidate_games g JOIN positions pos ON pos.game_id = g.game_id
      AND pos.ply = :ply_count WHERE g.game_day BETWEEN (SELECT iso_day FROM date_limits
      WHERE kind = 'start') AND (SELECT iso_day FROM date_limits WHERE kind = 'end')
    parameters:
      start_date: ${parameters.start_date}
      end_date: ${parameters.end_date}
      speed: ${parameters.time_control}
      side: ${parameters.player_color}
      player: ${steps.step_1.scalar_value}
      ply_count: ${parameters.ply_count}
    summary: Eligible games with stored opening SAN and post-ply evaluation
- id: step_4
  tool: query_database_readonly
  arguments:
    sql: SELECT :dataset_query || ' LIMIT :preview_count' AS preview_query
    parameters:
      dataset_query: ${steps.step_2.scalar_value}
    max_rows: 1
- id: step_5
  tool: orient_game_rows
  arguments:
    source_artifact: ${steps.step_3.artifact_id}
    player_identifier: ${steps.step_1.scalar_value}
    player_color: ${parameters.player_color}
    summary: Games oriented to the selected player's perspective
- id: step_6
  tool: query_workspace_readonly
  arguments:
    sources:
      oriented: ${steps.step_5.artifact_id}
    sql: ${steps.step_2.scalar_value}
    result_mode: workspace
- id: step_7
  tool: query_workspace_readonly
  arguments:
    sources:
      oriented: ${steps.step_5.artifact_id}
    sql: ${steps.step_4.scalar_value}
    parameters:
      preview_count: ${parameters.preview_row_count}
    result_mode: workspace
- id: step_8
  tool: export_workspace_dataset
  arguments:
    source_artifact: ${steps.step_6.artifact_id}
    output_filename: ${parameters.output_filename}
- id: step_9
  tool: present_scalar
  arguments:
    value: ${steps.step_6.row_count}
- id: step_10
  tool: present_table
  arguments:
    source_artifact: ${steps.step_7.artifact_id}
outputs:
- step_id: step_8
  effect_index: 0
  effect_type: file
- step_id: step_9
  effect_index: 0
  effect_type: screen
- step_id: step_10
  effect_index: 0
  effect_type: screen
\end{Verbatim}

\end{document}